\pdfoutput=1

\documentclass[preprint,12pt,authoryear,nopreprintline]{elsarticle}

\usepackage{lmodern}
\usepackage{amssymb}
\usepackage{amsmath}
\usepackage{booktabs}
\usepackage{multirow}
\usepackage{diagbox}
\usepackage[caption=false]{subfig}
\usepackage{microtype}
\usepackage{xurl}
\usepackage[hidelinks,hypertexnames=false]{hyperref}
\usepackage[capitalise,nameinlink]{cleveref}
\crefname{figure}{Figure}{Figures}
\crefname{equation}{Equation}{Equations}
\graphicspath{{figures/}}

\biboptions{sort}
\hypersetup{
    pdftitle={MOO: A Multi-view Oriented Observations Dataset for Viewpoint Analysis in Cattle Re-Identification},
    pdfauthor={William Grolleau, Achraf Chaouch, Astrid Sabourin, Guillaume Lapouge, Catherine Achard}
}

\begin{document}

    \begin{frontmatter}

        \title{MOO: A Multi-view Oriented Observations Dataset for Viewpoint Analysis in Cattle Re-Identification}

        \author[cea]{William Grolleau\corref{cor1}}
        \ead{william.grolleau@cea.fr}
        \author[cea]{Achraf Chaouch}
        \author[cea]{Astrid Sabourin}
        \author[cea]{Guillaume Lapouge\corref{cor1}}
        \ead{guillaume.lapouge@cea.fr}
        \author[isir]{Catherine Achard}
        \cortext[cor1]{Corresponding authors.}

        \affiliation[cea]{organization={Universit\'e Paris-Saclay, CEA, List},
            postcode={F-91120},
            city={Palaiseau},
            country={France}}
        \affiliation[isir]{organization={Sorbonne University, CNRS, Institute of Intelligent Systems and Robotics, ISIR},
            addressline={4 Place Jussieu},
            postcode={75005},
            city={Paris},
            country={France}}

        \begin{abstract}
            Animal re-identification (ReID) matches an individual across non-overlapping camera views despite changes in scene and viewpoint.
            It is a component of automated livestock monitoring, where identifying each animal is necessary to monitor its health and behavior over time.
            The available viewpoints depend on where cameras can be installed, so a camera network may have to combine very different viewpoints.
            Knowing which viewpoints give the best identification accuracy therefore helps decide where to place the cameras.
            However, existing cattle datasets do not provide the viewpoint coverage and annotations needed to measure how viewpoint affects identification.

            We introduce Multi-view Oriented Observations (MOO), a synthetic cattle ReID dataset in which viewpoint is precisely controlled without confounding factors.
            It contains $1{,}000$ cattle identities rendered from $128$ systematically sampled viewpoints with exact azimuth and elevation annotations.
            Cattle are particularly suitable for this study because their coat patterns are asymmetric and strongly dependent on viewpoint.

            We quantify the influence of viewpoint on identification accuracy and show that some viewpoints are much more informative than others.
            Elevation generalization on cattle is asymmetric, because a model trained on high elevations transfers downward better than a model trained on low elevations transfers upward.
            Below a $30^\circ$ elevation threshold, limited overlap between visible surfaces makes retrieval more difficult.
            Lateral views perform better than frontal and rear views, which we attribute to the larger amount of coat pattern they expose.
            Higher camera elevations and lateral views are therefore preferable in this controlled setting.

            We then define a benchmark in which matching top and side views remains difficult even under controlled synthetic conditions.
            Finally, MOO pretraining improves zero-shot retrieval on four real cattle datasets and supervised retrieval on two of them.
            Its viewpoint annotations let us select pretraining images that match the target dataset.
            MOO is publicly available at \url{https://github.com/TurtleSmoke/MOO}.
        \end{abstract}


    \end{frontmatter}


    \section{Introduction}
    \label{sec:introduction}

    Animal re-identification (ReID) matches an individual across non-overlapping camera views despite changes in pose and viewpoint.
    It is a core component of automated tracking for wildlife conservation and livestock management, where identifying each animal is necessary to monitor its health and behavior over time.

    For patterned animals, a viewpoint variation changes the perspective under which the coat markings are observed, so the same pattern can appear differently across images.
    We choose cattle as a case study of this problem because identifying them requires matching asymmetric coat markings, which are particularly sensitive to viewpoint changes.
    Cattle monitoring is also a well-studied application of animal ReID \citep{andrewVisualIdentificationIndividual2021a, gaoSelfSupervisionVideoIdentification2021b, xiaoCowIdentificationFreestall2022b, yuMultiCamCows2024MultiviewImage2024, zhaoCompactLossVisual2022, liIndividualDairyCow2021, shojaeipourAutomatedMuzzleDetection2021, bergaminiMultiviewsEmbeddingCattle2018a}.

\begin{figure}[htbp]
        \centering
        \includegraphics[width=0.85\linewidth]{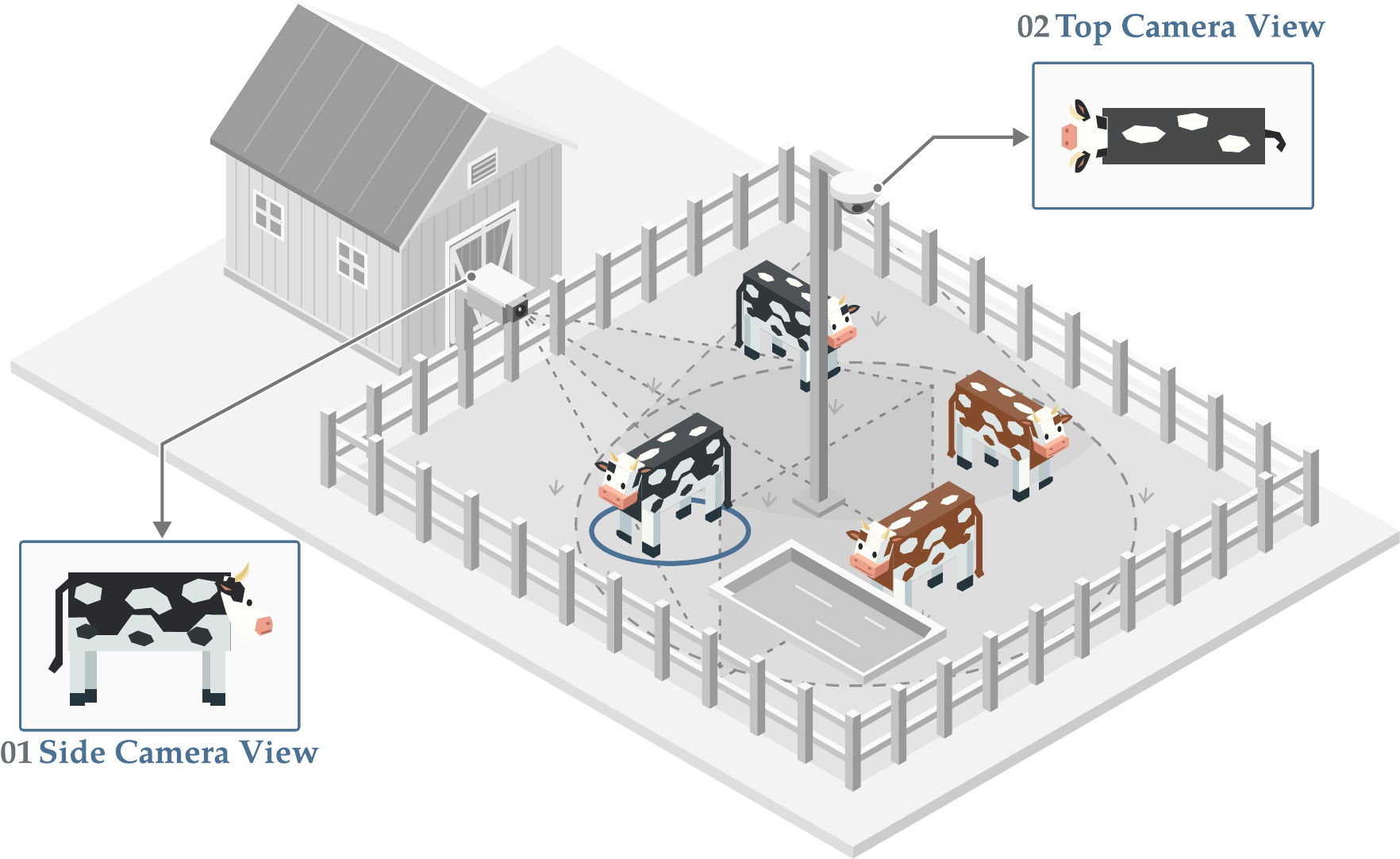}
        \caption{
            Camera placement changes the cattle markings available for identification.
            A ground camera captures a side view dominated by one flank (01), whereas an overhead camera captures the back and parts of both flanks (02).
            Matching these observations requires relating coat patterns seen under different projections and with different visible surfaces.
        }
        \label{fig:context-overview-moo}
    \end{figure}

    In practice, the available viewpoints depend on where cameras can be installed, as shown in \cref{fig:context-overview-moo} with a side camera (01) and a top camera (02).
    Their placement is usually constrained by the structure of enclosed spaces such as barns or by the terrain of natural environments, which can force a camera network to combine very different viewpoints.
    The gap is particularly large when ground and aerial cameras are combined, as in Aerial-Ground Re-Identification (AG-ReID).
    A ground camera observes the animal from roughly its own height and produces a side view, whereas an aerial camera looks down and produces a top view.
    A ceiling-mounted fisheye camera covers a large part of a barn, so a single image can contain both top and side views.
    An animal standing under the camera is seen from above, while one farther away is seen from a lower elevation.
    Knowing which viewpoints give the best identification accuracy therefore helps decide where to place the cameras.

    Existing animal ReID methods handle viewpoint changes by learning viewpoint-agnostic features, normalizing appearance geometrically, or combining several views (\cref{sec:related}).
    However, no existing work measures how much a given viewpoint change degrades identification, or when each of these strategies should be preferred.
    Studies on person ReID \citep{sunDissectingPersonReIdentification2019a} do not transfer directly to animals, since the asymmetric coat patterns of many species make their identity cues more sensitive to viewpoint.
    Measuring this effect requires exhaustive viewpoint coverage and precise angular annotations that currently do not exist in any animal ReID dataset.

    To fill this gap, we introduce the Multi-view Oriented Observations (MOO) dataset, a large-scale synthetic benchmark for cattle ReID with precise azimuth and elevation annotations.
    We first describe how the dataset is built and how its acquisition conditions are controlled, which provides the basis for studying viewpoint without confounding factors (\cref{sec:dataset}).
    We then measure the effects of elevation and azimuth (\cref{sec:atomic}).
    Based on this analysis, we define benchmark scenarios and evaluate how difficult they are for a baseline model (\cref{sec:benchmark}).
    Finally, we test whether features learned on MOO transfer to real cattle datasets (\cref{sec:transfer}).

    Our contributions are as follows:
    \begin{itemize}
        \item We present the MOO dataset, a large-scale synthetic benchmark for patterned animal ReID with precise azimuth and elevation annotations.
        \item We provide a systematic quantification of the impact of elevation and azimuth on patterned animal ReID, identifying a $30^\circ$ elevation angle as a key threshold for generalization and a gap between lateral views and frontal or rear views.
        \item We define a benchmark with three training configurations and five query-gallery scenarios in which viewpoint is the central difficulty.
        \item We demonstrate that MOO serves as a pretraining source for real-world cattle datasets in both zero-shot and supervised scenarios.
    \end{itemize}

    \section{Related Work}
    \label{sec:related}

    Most work on viewpoint changes in ReID focuses on people and does not always transfer to patterned animals.
    This section examines how animal ReID methods handle viewpoint changes and whether existing datasets support a systematic study of their effect.

    \subsection{Viewpoint Handling in Animal Re-Identification}
    \label{subsec:viewpoint-studies}

    For humans, some discriminative cues are viewpoint-invariant, such as clothing color, which is mostly unaffected by rotation.
    For animals, identity is carried by coat patterns, whose projection on the image plane changes significantly with viewpoint.
    Existing methods deal with this by learning viewpoint-agnostic features, normalizing appearance geometrically, or combining several views.

    \subsubsection{Learning viewpoint-agnostic features}

    The first strategy avoids modeling viewpoint explicitly and instead focuses on features that change little with it.
    Deep metric learning imposes this invariance through the loss by pulling images of the same individual together and pushing images of different individuals apart \citep{zhaoCompactLossVisual2022, zhaoMultiCenterAgentLoss2022}.
    Local representations pursue the same goal by restricting the information encoded by the model.
    They either combine global and local branches \citep{chenHolsteinCattleFace2022b}, aggregate whole-body patterns \citep{nepovinnykhSpeciesAgnosticPatternedAnimal2024, nepovinnykhReidentificationPatternedAnimals2025c} or focus on a discriminative region such as the muzzle, flank, or head \citep{shojaeipourAutomatedMuzzleDetection2021, dacLivestockIdentificationUsing2022}.

    Both approaches learn this invariance from the data alone, without any viewpoint supervision.
    In doing so, they may also discard viewpoint-specific cues that would still identify an individual if exploited correctly.

    \subsubsection{Geometric normalization}

    Instead of discarding viewpoint-specific cues, geometric normalization transforms the observed appearance into a canonical space.
    It uses the animal's geometry to map each observation into a common space that is independent of viewpoint and therefore comparable across views.
    For example, \citet{hiby2009tiger} use a 3D model to project a tiger's stripes onto a flat texture map, while \citet{algasovUnsupervisedPelagePattern2025} estimate surface normals to unwrap a seal's pelage pattern into a canonical 2D representation.

    This explicit normalization comes at a practical cost because it requires a 3D model of the individual or an estimate of its surface geometry, which is difficult to obtain for many species.

    \subsubsection{Multi-view aggregation}

    When geometry is unavailable, several observations can instead provide the information that may be missing from a single view.
    Multi-view methods combine these observations into a single representation that is more complete and more robust to viewpoint changes.
    They typically use multiple networks to learn from complementary regions \citep{li2024cattle} or from frontal and lateral views of a cow's face \citep{bergaminiMultiviewsEmbeddingCattle2018a}.

    Learning the relationships between views is complex and requires many training samples, which makes these methods less common than the preceding ones.
    Their use also depends on several observations or a multi-camera setup at inference time, which is not always practical.

    \subsubsection{Summary}
    These three strategies show that viewpoint is widely recognized as a major challenge in animal ReID.
    Each represents a different set of assumptions and trade-offs regarding supervision, model complexity, and acquisition constraints at inference time.
    Yet, none of these works measures how much a given viewpoint change actually affects ReID performance, nor the conditions under which each strategy becomes preferable in practice.
    Such an analysis exists for humans \citep{sunDissectingPersonReIdentification2019a}, but its findings do not transfer directly to animals because many species have asymmetric coat patterns, making their identity cues more sensitive to viewpoint.
    Providing this analysis for animals is therefore essential to guide end-users toward the right strategy, and it first requires a dataset that captures viewpoint variation in a controlled and exhaustive manner.

    \subsection{Datasets for Animal Re-Identification}
    \label{subsec:datasets}

    To perform a systematic analysis, a dataset must provide enough data coverage to isolate the effect of interest and enough annotation to measure it.
    In practice, an animal dataset collection is constrained by a trade-off between control and diversity.
    On the one hand, restricting animal motion or camera placement makes viewpoint easy to annotate, but reduces the generality of the dataset and the diversity of viewpoints.
    On the other hand, allowing animals to move freely is more realistic, but makes viewpoint annotation difficult.
    Having a good balance is key to studying the effect of viewpoint on ReID performance.

    \subsubsection{Cattle datasets}

    Cattle datasets sit at the controlled end of this trade-off.
    Because they are often collected in barns with few individuals, controlling camera placement or the animal's motion is an effective way to compensate for this limited data and simplify identification.
    The most common setup is either a top-down camera that can capture both flanks simultaneously \citep{andrewVisualIdentificationIndividual2021a, gaoSelfSupervisionVideoIdentification2021b, xiaoCowIdentificationFreestall2022b, yuMultiCamCows2024MultiviewImage2024} or lateral cameras that capture one flank at a time \citep{zhaoCompactLossVisual2022, liIndividualDairyCow2021}.
    Some datasets avoid whole-body correspondence by focusing on a small but distinctive region, such as the muzzle \citep{shojaeipourAutomatedMuzzleDetection2021} or the face from frontal and lateral views \citep{bergaminiMultiviewsEmbeddingCattle2018a}.

    Although these controlled setups enable high identification accuracy and simplify viewpoint annotation, they impose strong assumptions on the camera placement and the animal's motion.
    Restricting an animal's movement is both impractical in many monitoring scenarios and detrimental to its welfare.
    A recent dataset samples a wider range of orientations where the animal is free to move \citep{perneelDynamicMultiBehaviourOrientationInvariant2025}, but this sampling remains far from exhaustive.

    \subsubsection{Wildlife datasets}

    Wildlife datasets sit at the opposite end of this trade-off, since they try to capture animals in their natural habitat without any interference.
    To do so, they often use camera traps or unmanned aerial vehicles (UAVs) to record animals in their daily lives, which produces a wider range of viewpoints than controlled setups.
    As a result, they usually contain more diversity, as seen in datasets for tigers \citep{liATRWBenchmarkAmur2020}, zebras and giraffes \citep{parham2017animal}, and leopards \citep{bpct2022leopardid}.
    Some datasets and methods also restrict the viewpoint to a distinctive region, such as sea-turtle heads \citep{adamSeaTurtleID2022LongspanDataset2024a} or the belly spot patterns of manta rays \citep{moskvyakLearningLandmarkGuided2020}.

    Although these datasets cover more diverse viewpoints, that diversity comes with confounding factors such as occlusion, illumination changes, and background clutter.
    Because the animals move freely, the viewpoint distribution is uncontrolled and biased toward the most frequently observed orientations.
    The deeper obstacle is coverage, since an uncontrolled acquisition rarely observes every viewpoint of every individual, which a systematic study of the effect of viewpoint requires.

    \subsubsection{Synthetic datasets}

    Synthetic datasets could in principle capture a wide range of viewpoints while providing precise annotations, but at the cost of image realism.
    Synthetic data is already established for animal pose estimation \citep{Jiang2022BMVC, bonettoZebraPoseZebraDetection2024} and for segmentation \citep{muLearningSyntheticAnimals2020a, pengLearningPartSegmentation2024}, but it remains rare in ReID.
    To our knowledge, the only example is the dataset of \citet{picekCzechLynxDatasetIndividual2025}, which contains both real and synthetic images of lynxes, but provides no viewpoint annotation.

    \subsubsection{Summary}
    A systematic analysis of viewpoint in animal ReID requires a viewpoint coverage wide enough to isolate its effect and viewpoint annotations precise enough to measure it.
    None of these datasets satisfies both conditions.
    Cattle datasets offer control and annotation, but their viewpoint diversity is limited.
    Wildlife datasets offer diversity, but their viewpoint is uncontrolled and only coarsely annotated.
    Synthetic datasets could satisfy both, yet the only ReID example provides no viewpoint annotation.
    The datasets reviewed here do not provide the combined viewpoint coverage and annotations needed for the controlled analysis described in \cref{subsec:viewpoint-studies}.

    \Cref{tab:datasets} compares MOO with existing wildlife and cattle datasets.

    \begin{table}[htbp]
        \centering
        \resizebox{\linewidth}{!}{%
            \begin{tabular}{llrrlccl}
                \toprule

                \textbf{Species}
                & \textbf{Dataset}
                & \textbf{IDs}
                & \textbf{Images}
                & \textbf{View}
                & \textbf{Synth.}
                & \textbf{Uncorr.}
                & \textbf{Annotation} \\

                \midrule
                \multirow{6}{*}{Wildlife}

                & ATRW \citep{liATRWBenchmarkAmur2020}
                & 92 & 3{,}649 & Lateral & $\circ$ & $\bullet$ & \\

                & SealID \citep{nepovinnykhSealIDSaimaaRinged2022a}
                & 57 & 2{,}080 & Lateral & $\circ$ & $\bullet$ & \\

                & SeaTurtleID2022 \citep{adamSeaTurtleID2022LongspanDataset2024a}
                & 438 & 8{,}729 & Lateral & $\circ$ & $\bullet$ & Head orient. \\

                & Nyala Data \citep{dlaminiAutomatedIdentificationIndividuals2020b}
                & 237 & 1{,}942 & Lateral & $\circ$ & $\bullet$ & Left vs Right \\

                & LeopardID2022 \citep{bpct2022leopardid}
                & 430 & 6{,}795 & Lateral & $\circ$ & $\bullet$ & Left vs Right \\

                & CzechLynx \citep{picekCzechLynxDatasetIndividual2025}
                & 319 & 39k & Lateral & $\bullet$ & $\bullet$ & \\

                \midrule
                \multirow{10}{*}{Cattle}

                & Multi-views Embedding \citep{bergaminiMultiviewsEmbeddingCattle2018a}
                & 439 & 17{,}802 & Lateral & $\circ$ & $\circ$ & Front vs Side \\

                & OpenCows2020 \citep{andrewVisualIdentificationIndividual2021a}
                & 46 & 4{,}736 & Top & $\circ$ & $\circ$ & \\

                & Cows2021 \citep{gaoSelfSupervisionVideoIdentification2021b}
                & 182 & 13{,}784 & Top & $\circ$ & $\circ$ & \\

                & MuzzleCows \citep{shojaeipourAutomatedMuzzleDetection2021}
                & 300 & 2{,}900 & Face & $\circ$ & $\bullet$ & \\

                & Dairy Cows 2021 \citep{liIndividualDairyCow2021}
                & 13 & 1{,}485 & Lateral & $\circ$ & $\bullet$ & \\

                & MVCAID100 \citep{zhaoCompactLossVisual2022}
                & 100 & 4{,}073 & Lateral & $\circ$ & $\circ$ & \\

                & Cows Stall Barn \citep{xiaoCowIdentificationFreestall2022b}
                & 48 & 8{,}640 & Top & $\circ$ & $\circ$ & \\

                & MultiCamCows2024 \citep{yuMultiCamCows2024MultiviewImage2024}
                & 90 & 101{,}329 & Top & $\circ$ & $\circ$ & \\

                & CoBRA ReID \citep{perneelDynamicMultiBehaviourOrientationInvariant2025}
                & 48 & 11{,}438 & Top & $\circ$ & $\circ$ & Az. \\

                & \textbf{MOO}
                & \textbf{1{,}000} & \textbf{128{,}000} & \textbf{All} & $\bullet$ & $\bullet$ & \textbf{Az. \& El.} \\

                \bottomrule
            \end{tabular}%
        }
        \caption{
            Animal ReID datasets compared by scale, viewpoint coverage, and annotation.
            IDs gives the number of individuals.
            Synth. indicates synthetic images and Uncorr. indicates independent frames rather than video sequences.
            Filled circles mean yes and open circles mean no.
            Az. and El. denote azimuth and elevation labels. Named views denote categorical annotations, and blank cells indicate no reported viewpoint annotation.
        }
        \label{tab:datasets}
    \end{table}

    Compared with these datasets, MOO has more individuals and images, a larger viewpoint coverage, and more precise viewpoint annotations.
    Furthermore, its controlled setting allows us to isolate the effect of viewpoint on ReID performance, which is not possible with existing datasets.

    \section{The MOO Dataset}
    \label{sec:dataset}

    To isolate the effect of viewpoint in patterned animal ReID, MOO prioritizes control over photorealism.
    While photorealistic images would introduce the confounding factors found in real data, control is what lets us attribute a performance change to the viewpoint alone.

    \subsection{Building the Individuals}
    \label{subsec:individuals}

\begin{figure}[htbp]
        \centering
        \includegraphics[width=0.85\linewidth]{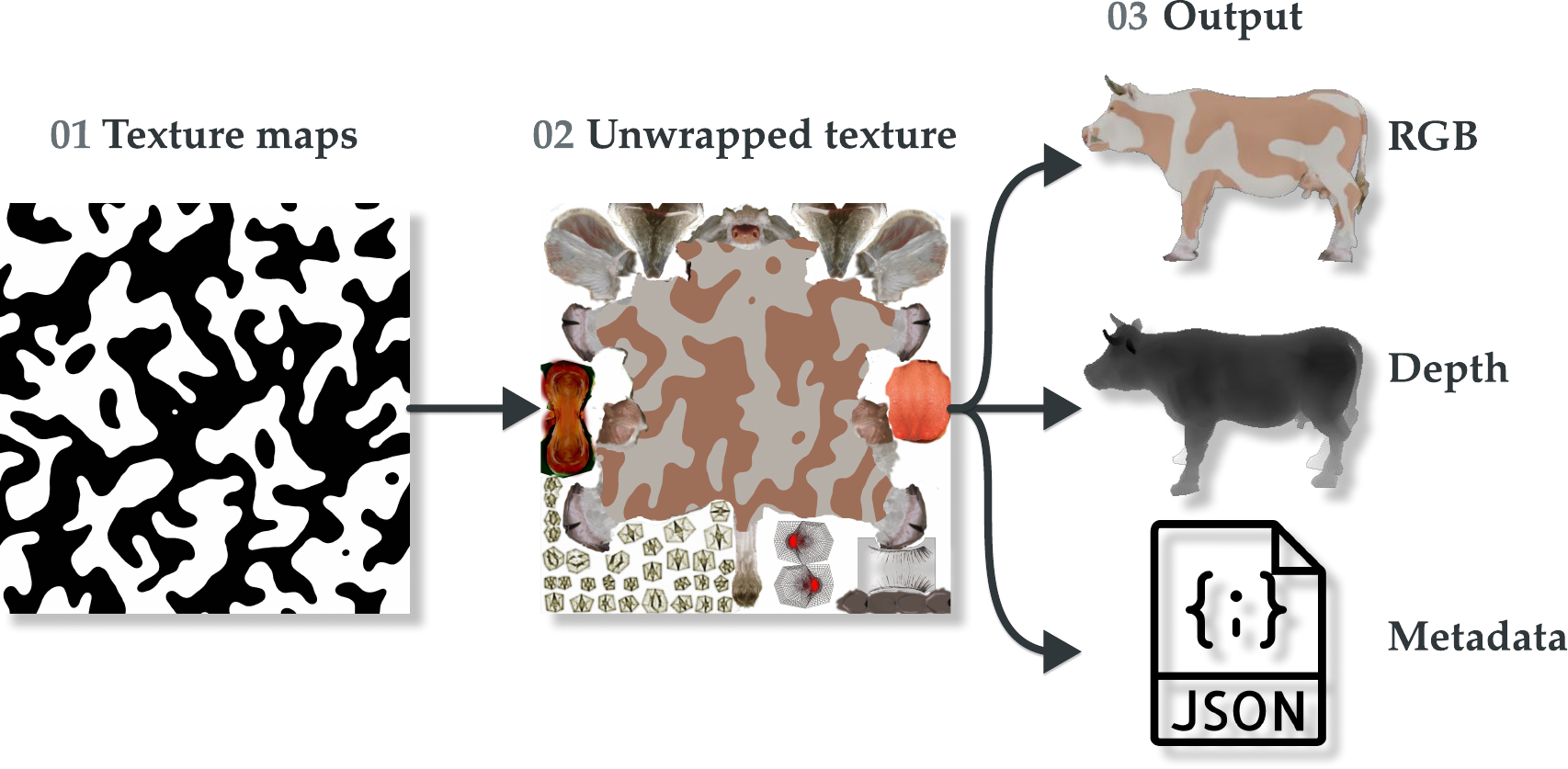}
        \caption{
            Generation of MOO identities and observations.
            Procedural coat textures (01) are mapped on the UV surface of a cow mesh (02).
            This mesh is then rendered from different camera positions to produce RGB images, depth maps, and metadata (03).
        }
        \label{fig:pipeline}
    \end{figure}

    We designed the generation pipeline in \cref{fig:pipeline} to create distinct cattle identities while controlling every factor other than coat pattern and viewpoint.
    Our pipeline combines procedural texture synthesis (\cref{fig:pipeline}, 01), texture mapping (02), and controlled rendering (03) to produce the animals, their images, and the associated metadata.

    \paragraph{Cattle mesh}
    We use the same realistic 3D mesh \citep{turbosquid_cow} in a neutral standing pose for every cow.
    This fixed geometry removes pose and shape variations that could become discriminative and confound the viewpoint analysis.
    The mesh is also UV-unwrapped to produce a 2D representation of its surface, which is used to map the texture onto the 3D geometry.

    \paragraph{Texture generation}
    To create distinct identities on this fixed mesh, we generate coat patterns with a procedural synthesis framework \citep{dumery_cow_tex}.
    This framework generates cumulative noise patterns that are thresholded to produce black-and-white texture maps, which we recolor to brown and white to reproduce common cattle coats.
    The discriminative cues are therefore the density, size, shape and spatial arrangement of the blobs across the coat.

    \paragraph{Texture mapping}
    We map each generated 2D texture onto the 3D mesh using its UV coordinates, which associate locations in the texture with the 3D surface.
    We leave the hooves, eyes, udders, and tail untextured, since real cows rarely carry markings there.

    \subsection{Rendering the Images}
    \label{subsec:rendering-the-images}

    With each individual built, we can render it in Blender \citep{blender} under near-uniform lighting and vary the viewpoint between images.
    This lighting reveals the mesh geometry without casting strong shadows that a model could exploit.
    For the same reason, images have no rendered background, and a foreground mask isolates the cow in each one.
    Together, these choices prevent the scene from carrying identity information that may bias the model \citep{picekAnimalIdentificationIndependent2024a, tianEliminatingBackgroundbiasRobust2018, yuAddressingElephantRoom2024}.

\begin{figure}[htbp]
        \centering
        \includegraphics[width=0.85\linewidth]{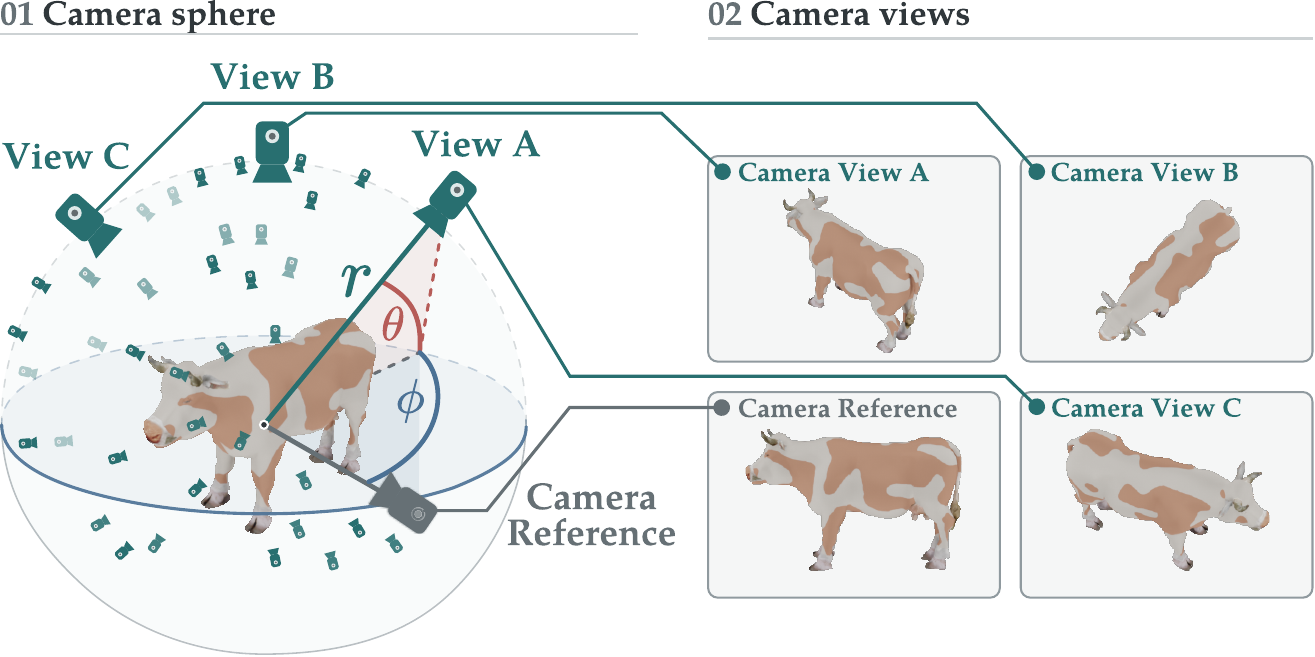}
        \caption{
            Spherical coordinate system in MOO, with azimuth $\phi$, elevation $\theta$, and radius $r$ illustrated for Camera A (01).
            The elevation $\theta$ measures Camera A's angle above the equatorial plane, while the azimuth $\phi$ measures its horizontal rotation from the reference direction.
            The reference view and views A, B, and C (02) illustrate how changing camera position changes the projection of the cow's coat.
        }
        \label{fig:camera-geometry}
    \end{figure}

    The camera position is set by an azimuth angle $\phi$ and an elevation angle $\theta$.
    The camera stays at a fixed distance $r$ from the cow, so the whole animal remains in frame at the same scale for every angle.
    \Cref{fig:camera-geometry} illustrates this coordinate convention (01) and the renders obtained from four camera positions (02).

    We sample $16$ azimuths uniformly over $\phi \in [0^\circ, 360^\circ)$ and $8$ elevations over $\theta \in [-20^\circ, 85^\circ]$, where $(0^\circ, 0^\circ)$ is the lateral view of a cow facing left.
    The full azimuth range supports a systematic analysis, while the elevation range spans views from slightly below the body, as with low cameras in barns or corridors, to top-down views from ceiling cameras or drones.

    Finally, random jitters of $\pm{10}^\circ$ in azimuth and $\pm{5}^\circ$ in elevation prevent the discrete sampling grid from becoming a shortcut for the model.
    They also simulate continuous variation and extend the effective elevation range to $[-25^\circ, 90^\circ]$.
    The resulting dataset contains $128$ images of each of the $1{,}000$ individuals, for a total of $128{,}000$ images at $512 \times 512$ pixels.
    \Cref{fig:partitions} illustrates the partitions defined in \cref{subsec:atomic-splits} with an example of camera shift, and \cref{fig:dataset-viewpoint} shows one individual on the $8 \times 8$ grid of partition centers.
    The example renders of both figures omit the $\pm{10}^\circ$ azimuth and $\pm{5}^\circ$ elevation jitters for clarity.

    \subsection{Metadata}
    \label{subsec:metadata}

    Alongside each RGB image, the rendering process produces a segmentation mask, a depth map, and the metadata needed to select and analyze viewpoints.
    While the mask isolates the animal, the depth map exposes its geometry, which can be used as an additional cue for ReID itself or for geometric tasks defined on the same samples.

    Each image records its exact azimuth $\phi \in [0^\circ, 360^\circ)$ and elevation $\theta \in [-25^\circ, 90^\circ]$ after jitter.
    Semantic labels also give a coarse direction, such as ``front-right'' or ``top-front-right'', among $8$ azimuth sectors of $45^\circ$.
    Spatial metadata includes the camera position $(x, y, z)$ in world coordinates and the homogeneous transformation matrix containing its rotation and translation.
    It is completed by the intrinsic and extrinsic matrices of the camera used for rendering.

    Together, the semantic labels and precise camera parameters allow both deployment-oriented protocols and a fine-grained viewpoint analysis.

    \subsection{Training and Test Identities}
    \label{subsec:splits}

    We divide the dataset into $500$ identities for training and $500$ for testing, with no identity shared between them.
    This identity split is common to the viewpoint analysis and the benchmark.
    Within it, we select images by angle to form the atomic partitions of \cref{subsec:atomic-splits} and the deployment-oriented groups of \cref{subsec:primary-split}.
    Query and gallery sets are always disjoint.
    In the atomic analysis, each test partition is split into a query set and a gallery set, so query and gallery images always come from the same partition.
    The deployment-oriented groups split the elevation range at the $30^\circ$ threshold identified in \cref{subsec:elevation}.
    Side contains images below $30^\circ$, Top contains images at or above $30^\circ$, and Top-Side contains both Top and Side images.
    For the Side, Top and Top-Side groups, the test images of each identity are divided in half, one half for the query set and the other half for the gallery set.

    \section{Experimental Setup}
    \label{sec:setup}

    \subsection{Re-Identification Baseline}
    \label{subsec:baseline}

    Following common practice among state-of-the-art ReID methods, we use the Vision Transformer (ViT) Bag of Tricks (ViT-BoT) architecture \citep{he2021transreid} without overlapping patches as the backbone of our baseline.
    The ViT divides the image $I_i$ into non-overlapping patches of $16\times16$ pixels and projects each one into a token.
    Following \citet{dosovitskiy2021image}, a learnable \emph{class token} is placed in front of these patch tokens, and a learnable positional embedding is added to every token.
    After the final transformer layer, only the class token is kept as the identity feature $z_i$ used to compare images.
    The backbone is initialized from a ViT \citep{dosovitskiy2021image} pretrained for image classification on ImageNet-21k \citep{deng2009imagenet}, and then fine-tuned on the ReID dataset.

    Training the baseline requires a training set $\{(I_i, y_i)\}$ of images $I_i$, each associated with an identity label $y_i$ among the $C$ training identities.
    The label $y_i$ supervises an identity loss and a triplet loss, which together shape the identity feature $z_i$.
    The Batch Normalization Neck (BNNeck) introduced by \citet{luo2019bag} creates a second version $h_i = \operatorname{BN}(z_i)$ of the class token, where $\operatorname{BN}$ denotes batch normalization.
    A linear classifier converts $h_i$ into one score for each of the $C$ training identities, and a softmax turns these scores into probabilities $p_i(y)$.
    For a mini-batch of $B$ images, cross-entropy gives
    \begin{equation}
        \mathcal{L}_{\text{id}}
        = -\frac{1}{B}\sum_{i=1}^{B}\log p_i(y_i).
        \label{eq:identity-loss}
    \end{equation}
    Following \citet{hermans2017defense}, the triplet loss is applied to the identity feature $z_i$ with $PK$ sampling, choosing $P$ identities and $K$ images per identity, giving a batch size of $B=PK$ images.
    Distances between features are measured by the Euclidean distance $D(u,v)=\lVert u-v\rVert_2$.
    For each anchor $z_a$ in the batch, batch-hard mining selects the farthest feature $z_a^+$ from another image of the same identity and the nearest feature $z_a^-$ with a different identity.
    Following TransReID \citep{he2021transreid}, we use its soft-margin form
    \begin{equation}
        \mathcal{L}_{\text{tri}}
        = \frac{1}{B}\sum_{a=1}^{B}
        \log\left(1+\exp\left(D(z_a,z_a^+)-D(z_a,z_a^-)\right)\right).
        \label{eq:triplet}
    \end{equation}
    The two terms are combined as $\mathcal{L}=\mathcal{L}_{\text{id}} + \lambda \mathcal{L}_{\text{tri}}$, with $\lambda=1$.
    At inference, the classifier, BNNeck, and $h_i$ are discarded.

    \subsection{Implementation Details}
    \label{subsec:implementation}

    Images are resized to $H \times W = 256 \times 256$ pixels, and training lasts $120$ epochs.
    We use stochastic gradient descent (SGD) with momentum $0.9$, weight decay $10^{-4}$, and a base learning rate of $0.008$ under cosine decay, after a linear warmup over the first $5$ epochs.
    We use a batch of size $128$, retaining $4$ images per identity, so each batch contains $32$ different identities.
    For the MOO experiments, the linear identity classifier has $C=500$ outputs, one per training identity, and for the supervised transfer, $C$ is the number of training identities in the target dataset.
    The training augmentations used here are random cropping and horizontal flipping.
    We keep horizontal flipping for consistency with common practice and examine its interaction with azimuth in \cref{subsec:hflip}.
    At test time, each image is represented by its feature $z_i$, and for each query, gallery images are ranked by increasing Euclidean distance to the query feature.
    We report the mean Average Precision (mAP) and the rank-1 accuracy computed from this ranking.

    \section{Isolating Elevation and Azimuth}
    \label{sec:atomic}

    To distinguish the effects of camera elevation and azimuth around the animal, we first compare elevation partitions that each contain every azimuth, then azimuth partitions that each contain every elevation.
    This separates the effect of elevation from that of azimuth under the controlled appearance conditions of MOO.

    \subsection{Atomic Viewpoint Partitions}
    \label{subsec:atomic-splits}

\begin{figure}[htbp]
        \centering
        \includegraphics[width=0.6\linewidth]{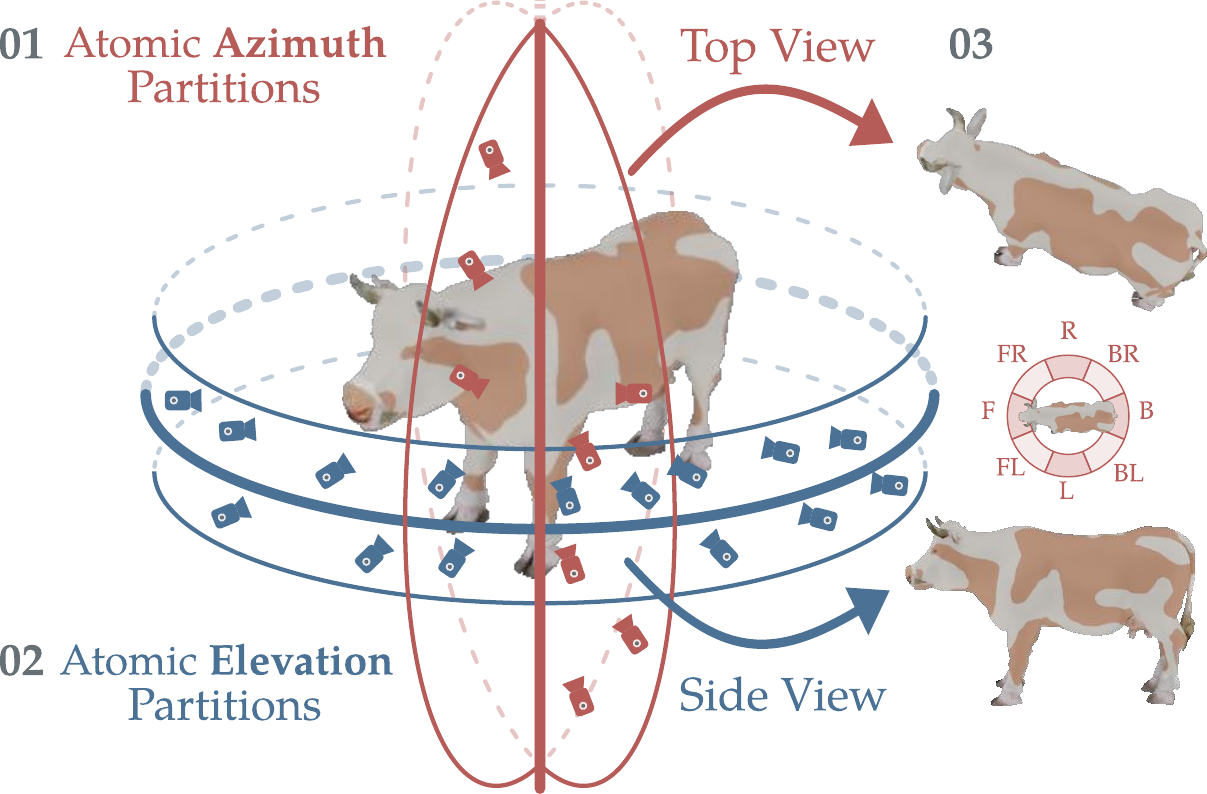}
        \caption{
            Schematic viewpoint partitions used to separate elevation and azimuth effects in MOO.
            An azimuth partition (01, red) restricts the side of observation while retaining all elevations, and an elevation partition (02, blue) restricts camera height while retaining all azimuths.
            The bands illustrate angular variation within a partition, and the example renders show top and side observations.
            The inset (03) shows the $8$ azimuth partitions seen from above, labeled front (F), front-right (FR), right (R), back-right (BR), back (B), back-left (BL), left (L), and front-left (FL).
        }
        \label{fig:partitions}
    \end{figure}

    \paragraph{Azimuth partitions}
    The $16$ rendered azimuths form $8$ semantic partitions separated by $45^\circ$, covering left, front-left, front, front-right, right, back-right, back, and back-left.
    Left and right name the flank of the animal that faces the camera, so the reference view $(0^\circ,0^\circ)$ shows the left flank.
    Each azimuth partition contains all $8$ sampled elevations, including their jitter.
    Restricting azimuth does not restrict the camera to a specific height, but only to a specific side of the animal.
    In \cref{fig:partitions}, the red line (01) marks one azimuth partition center, the surrounding band shows its jitter across all elevations, and the inset (03) shows the $8$ azimuth partitions seen from above.

    \paragraph{Elevation partitions}
    The $8$ rendered elevations define disjoint partitions centered at $\theta \in \{-20^\circ, -5^\circ, 10^\circ, 25^\circ, 40^\circ, 55^\circ, 70^\circ, 85^\circ\}$, each with a jitter of $\pm 5^\circ$.
    Each elevation partition contains all $16$ sampled azimuths, including their jitter.
    Restricting elevation does not restrict the camera to one side of the animal, but only to a specific height above the ground.
    In \cref{fig:partitions}, the blue line (02) marks one elevation partition center, and the surrounding band shows its jitter across all azimuths.

\begin{figure}[htbp]
        \centering
        \includegraphics[width=0.6\linewidth]{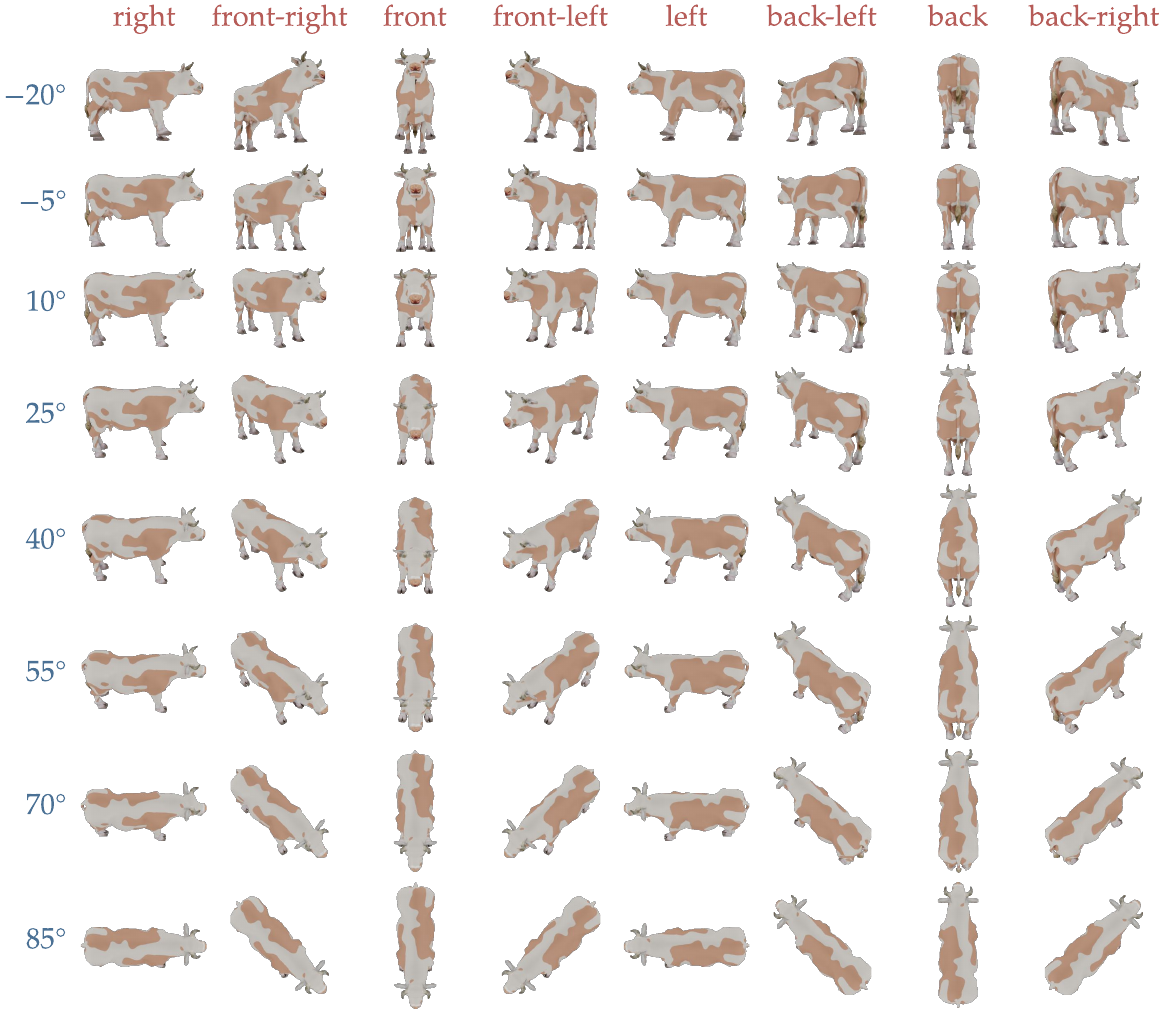}
        \caption{
            One MOO individual rendered at the center of each viewpoint partition.
            Rows are the $8$ elevation partitions and columns are the $8$ azimuth partitions.
            This illustrative example omits the jitter applied to the dataset images, so each render sits exactly at its partition center.
        }
        \label{fig:dataset-viewpoint}
    \end{figure}

    \Cref{fig:dataset-viewpoint} shows one individual rendered at the center of every elevation and azimuth partition.

    \subsection{The Impact of Elevation}
    \label{subsec:elevation}

    We begin with elevation, using the $8$ partitions previously defined and retaining for each the full azimuth range in training, query, and gallery sets.
    A training configuration is denoted by $\theta^\circ$, where $\theta$ is the central elevation of its partition.

    \subsubsection{Training on a single elevation}

    To establish how sensitive a model is to its training elevation, we train one model on each partition and test it on all eight.
    Within each test, query and gallery images come from the same elevation.
    The query-gallery relation is therefore fixed, leaving the gap between training and test elevations as the only varying factor.

\begin{figure}[htbp]
        \centering
        \includegraphics[width=0.5\linewidth]{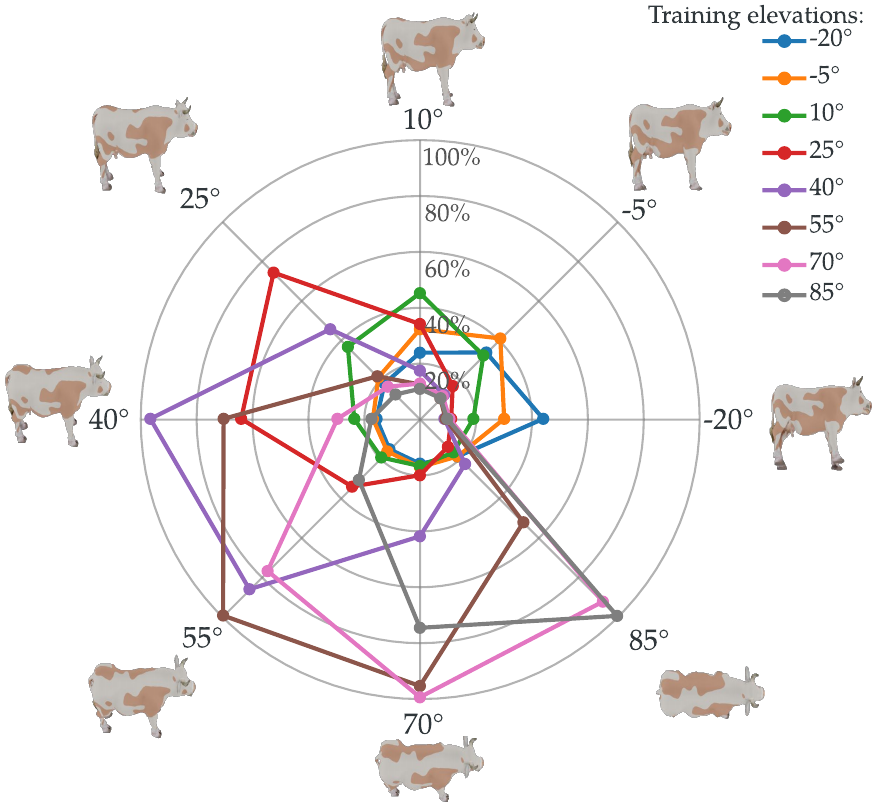}
        \caption{
            Generalization from one training elevation to the eight MOO test elevations.
            Each colored curve is a separately trained model. Angular positions identify the test elevation and radial distance gives mAP in percent.
            At each position, query and gallery share the same elevation partition, and training and evaluation retain all azimuths.
        }
        \label{fig:elev-single}
    \end{figure}

    \Cref{fig:elev-single} reports the result.
    As expected, each model performs best on its training partition, and performance decreases as the elevation gap widens.
    More importantly, this decrease is asymmetric because training below the inference elevation causes a much larger loss than training above it.
    Thus, models trained at high elevations generalize downward better than low-elevation models generalize upward.

    A second trend accompanies this asymmetry, as higher test elevations also perform better in absolute terms, which we attribute to their greater visibility of both flanks.
    Single-elevation experts perform substantially better in the partitions centered at $40^\circ$ and above.
    We therefore use $30^\circ$, the upper edge of the lower group, to define the benchmark split.
    Above it, enough of both flanks is visible for matching, while below it self-occlusion does not leave enough common cues between images taken from opposite azimuths.

    \subsubsection{Training on cumulative elevation ranges}

\begin{figure}[htbp]
        \centering
        \renewcommand{\thesubfigure}{0\arabic{subfigure}}
        \subfloat[Descending order, from the highest elevation downward.]{%
            \includegraphics[width=0.47\linewidth]{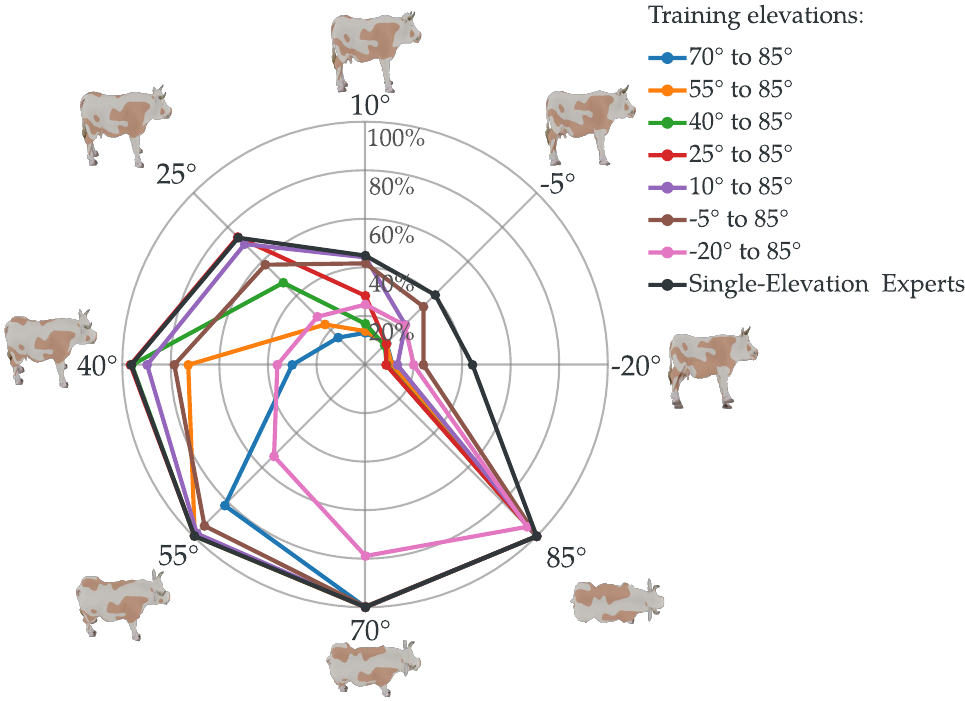}%
            \label{fig:cumul-down}%
        }
        \hfill
        \subfloat[Ascending order, from the lowest elevation upward.]{%
            \includegraphics[width=0.47\linewidth]{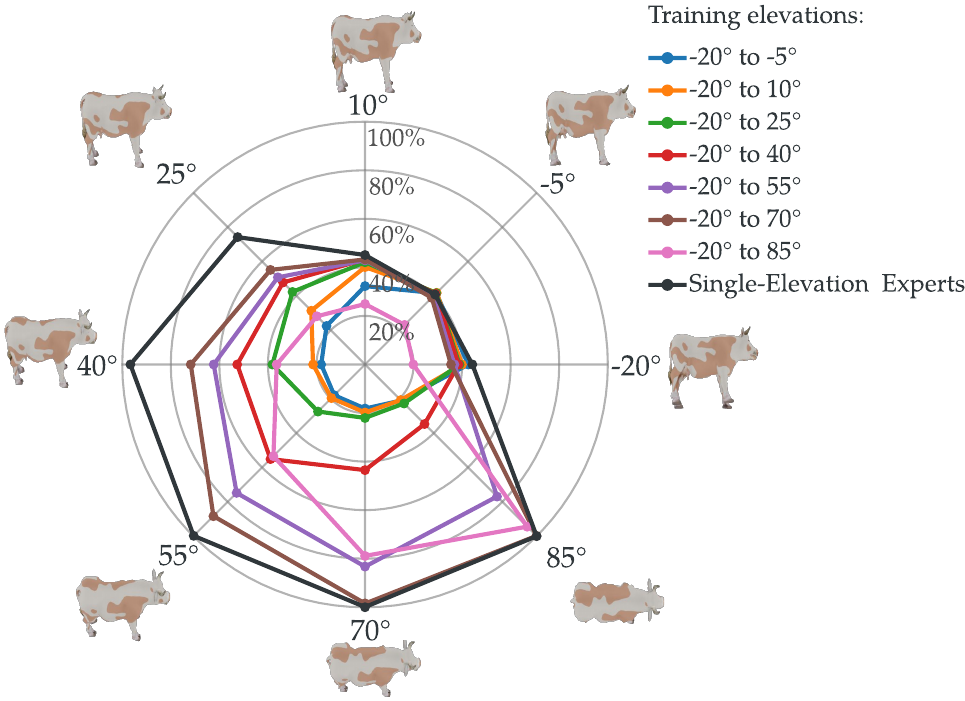}%
            \label{fig:cumul-up}%
        }
        \caption{
            mAP on MOO as training expands from high to low elevations (01) or low to high elevations (02).
            Each colored curve uses the inclusive range of elevation partitions named in the legend. Angular positions give test elevation and radial distance gives mAP in percent.
            Query and gallery share the test partition, with all azimuths retained.
            The black curve labeled ``Single-Elevation Experts'' combines the measured results of the separately trained single-elevation models and is an empirical specialization reference.
        }
        \label{fig:cumulative}
    \end{figure}

    The single-elevation experiment shows the benefit of specialization, but a deployed model must usually cover more than one camera height.
    We therefore train on cumulative elevation ranges, either from the highest elevation downward (\cref{fig:cumulative}, 01) or from the lowest upward (02).
    The notation $\theta_n^\circ$ to $\theta_m^\circ$ gives the first and last partitions in each range.

    The wider training ranges improve coverage of test elevations, although their performance remains below the corresponding single-elevation experts at many test elevations.
    Moreover, the descending ranges (\cref{fig:cumulative}, 01) confirm the $30^\circ$ threshold because once lower elevations enter the training range, performance on higher views noticeably decreases.

    If the model learned to use the animal's geometry to project observed patterns into a canonical space, this could help it retain the accuracy of single-elevation experts across elevations.
    Instead, the lack of generalization suggests that adding viewpoints alone does not make this baseline learn such a geometric representation.

    \subsubsection{Training on all elevations}

\begin{figure}[htbp]
        \centering
        \includegraphics[width=0.6\linewidth]{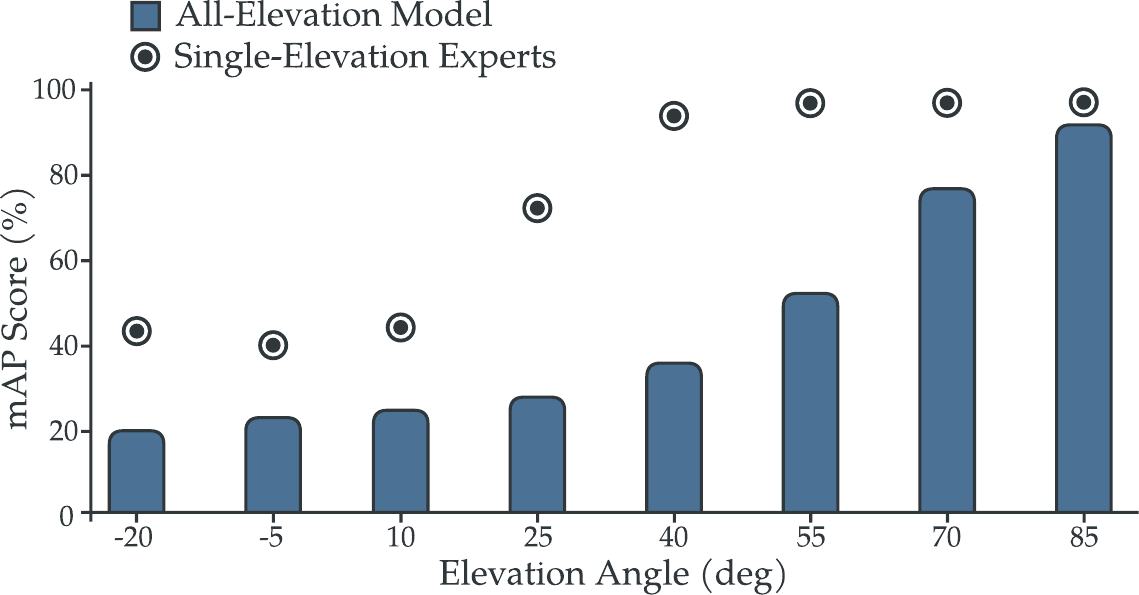}
        \caption{
            An all-elevation model (blue) compared with a separate single-elevation expert at each test elevation (black markers) on MOO.
            The horizontal axis gives elevation and the vertical axis gives mAP in percent.
            Query and gallery use the same elevation partition and retain all azimuths, exposing the loss of specialization when one model must cover every elevation.
        }
        \label{fig:expert}
    \end{figure}

    To confirm this limitation, we compare a model trained on all elevations with the single-elevation expert of each elevation.
    As \cref{fig:expert} shows, the all-elevation model falls short of these experts.
    It still generalizes better across viewpoints than a model trained on a single elevation, as the cumulative results of \cref{fig:cumulative} show.
    Covering every elevation thus helps generalization but does not match a specialized expert at each elevation.

    \subsection{The Impact of Azimuth}
    \label{subsec:azimuth}

    To study azimuth, we train models on the four cardinal partitions (left, right, front, and back) and evaluate them across all eight azimuth partitions defined in \cref{subsec:atomic-splits}.
    The all-directions model in \cref{fig:azimuth} provides a reference trained across the complete azimuth range.
    Each azimuth partition contains all eight elevations in training, query, and gallery sets.
    As before, each evaluation uses query and gallery images from the same partition, so the analysis isolates the relation between training and test azimuths.

    \subsubsection{Lateral versus frontal and rear views}

\begin{figure}[htbp]
        \centering
        \includegraphics[width=0.6\linewidth]{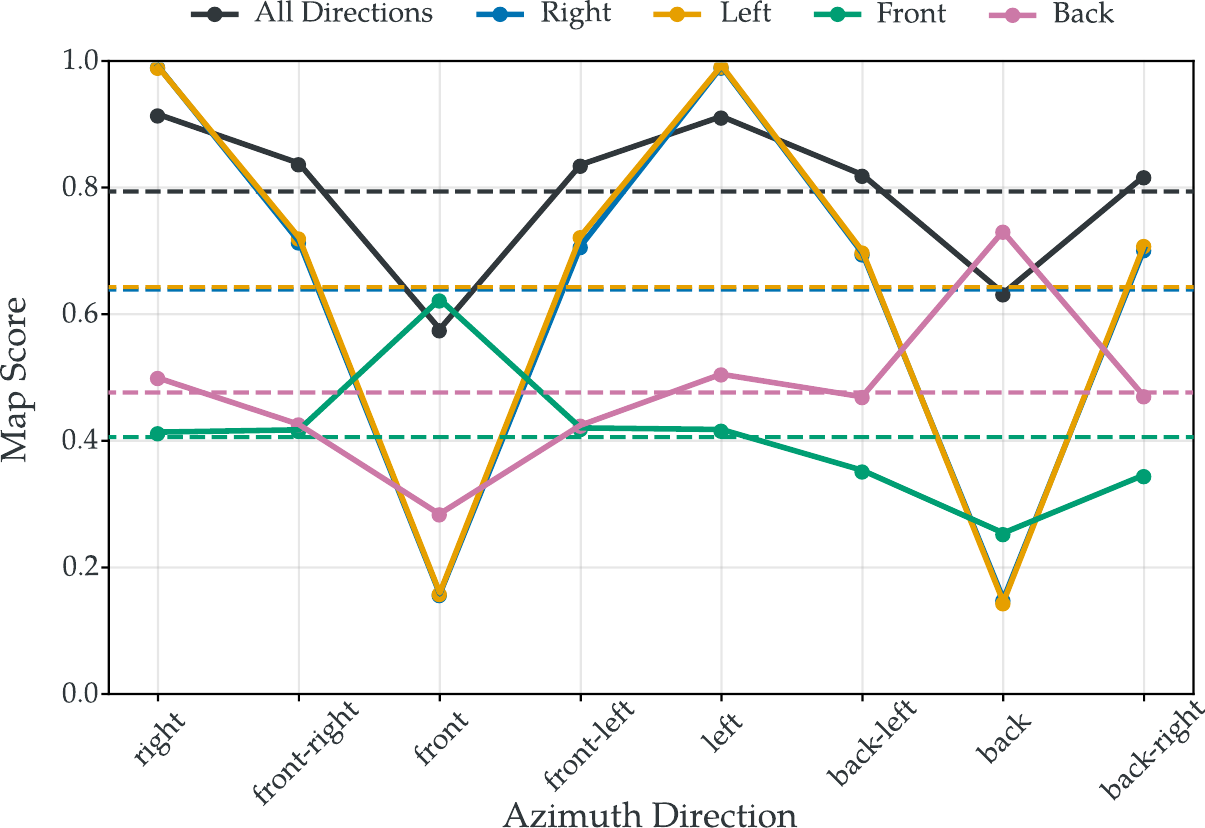}
        \caption{
            mAP across eight test azimuth partitions for models trained on right, left, front, back, or all directions.
            Solid curves show performance per test partition, and dashed horizontal lines show each model's mean across partitions.
            ``All Directions'' denotes training across the full azimuth range, unlike the four models trained on one partition each.
            Query and gallery share the test azimuth partition, with all elevations retained.
            The vertical axis uses a $0$--$1$ scale.
        }
        \label{fig:azimuth}
    \end{figure}

    The results of these models are reported in \cref{tab:azimuth} and \cref{fig:azimuth}.
    They show a clear gap between lateral views (left and right) and frontal and rear views.
    Training and testing on lateral views yields at least $0.98$ mAP, while the same lateral models reach at most $0.17$ on frontal and rear views.
    Averaged over the eight test partitions, lateral training reaches about $0.64$ mAP, compared with about $0.44$ for frontal and rear training.

    Query and gallery use the same azimuth partition, but how much visible surface two of their views share still depends on the partition.
    Within a lateral partition, a change in azimuth moves the camera along the same flank, so each view shows mostly the same coat region from a slightly different angle.
    Within a frontal or rear partition, the same change in azimuth can expose the left side of the animal in one view and the right side in another, so two views of the same cow share less visible surface.
    The amount of pattern on the visible surface also matters.
    A lateral view shows one full flank, which carries most of the coat markings.
    In contrast, a frontal or rear view mostly captures the head, chest or hindquarters, where the texture carries fewer markings and excludes regions such as the udders and tail (\cref{sec:dataset}).

    \begin{table}[htbp]
        \centering
        \setlength{\aboverulesep}{0pt}
        \setlength{\belowrulesep}{0pt}
        \renewcommand{\arraystretch}{1.2}
        \begin{tabular}{l|cccc|c}
            \toprule
            \diagbox{\textbf{Train}}{\textbf{Test}} & Right         & Left          & Front         & Back          & Average       \\
            \midrule
            Right                                   & \textbf{0.98} & 0.99          & 0.16          & 0.15          & 0.63          \\
            Left                                    & 0.99          & \textbf{0.99} & 0.17          & 0.16          & \textbf{0.65} \\
            Front                                   & 0.42          & 0.42          & \textbf{0.62} & 0.24          & 0.40          \\
            Back                                    & 0.50          & 0.51          & 0.29          & \textbf{0.75} & 0.48          \\
            \bottomrule
        \end{tabular}
        \caption{
            Azimuth generalization on MOO, with training partitions in rows and test partitions in columns.
            Query and gallery share the test azimuth partition, and all partitions retain every elevation.
            mAP is reported on a $0$--$1$ scale.
            Average covers all eight test azimuth partitions, including the intermediate directions absent from the table.
            Bold marks same-partition evaluations and the highest average.
        }
        \label{tab:azimuth}
    \end{table}

    \subsubsection{Horizontal flipping and lateral views}
    \label{subsec:hflip}

    Because cattle coats are asymmetric, horizontal flipping may interact with the azimuth partitions in a non-trivial way.
    Indeed, a flipped left flank has the orientation of a right view but shows a pattern that no real right view of the same cow would have.
    A model trained on one flank also sees the other orientation during training, which may explain why the left and right rows of \cref{tab:azimuth} are almost identical.
    To measure this effect, we repeat the azimuth training with and without flipping, and compare both settings in \cref{tab:hflip}.
    Without flipping, a left-trained model reaches $0.99$ mAP on left views and $0.96$ on right views, and a right-trained model is symmetric.
    When flipping is removed, the results are very similar, so the horizontal flipping augmentation seems to have no effect on lateral training.

    \begin{table}[htbp]
        \centering
        \begin{tabular}{llcccc}
            \toprule
            \textbf{Train}         & \textbf{HFlip} & Right & Left & Front & Back \\
            \midrule
            \multirow{2}{*}{Right} & \checkmark     & 0.98  & 0.99 & 0.16  & 0.15 \\
            &                & 0.99  & 0.95 & 0.17  & 0.16 \\
            \midrule
            \multirow{2}{*}{Left}  & \checkmark     & 0.99  & 0.99 & 0.17  & 0.16 \\
            &                & 0.96  & 0.99 & 0.17  & 0.16 \\
            \bottomrule
        \end{tabular}
        \caption{
            Azimuth generalization on MOO for lateral training with and without horizontal flipping.
            Query and gallery share the test azimuth partition, and all partitions retain every elevation.
            mAP is reported on a $0$--$1$ scale.
        }
        \label{tab:hflip}
    \end{table}

    \subsection{Summary}
    \label{subsec:atomic-summary}

    On cattle, elevation has an asymmetric effect, since models trained at high elevations generalize downward better than low-elevation models generalize upward.
    A threshold near $30^\circ$ separates views that reveal enough of both flanks for reliable matching from views where self-occlusion leaves few common surfaces.
    For azimuth, we observed a gap between lateral views and frontal or rear views, which we attribute to the amount of visible surface shared between query and gallery images and to the amount of coat pattern this surface carries.
    Training on several viewpoints improves coverage but does not recover the accuracy of a specialized model at each viewpoint, possibly because this baseline does not learn a geometric representation of the animal.
    These results motivate the benchmark scenarios of the next section, which aim to evaluate ReID methods on their ability to handle viewpoint changes.

    \section{The MOO Benchmark}
    \label{sec:benchmark}

    The elevation analysis of \cref{subsec:elevation} shows that elevation is a major factor of difficulty, with low and high elevations behaving as two distinct matching problems that a model does not easily bridge.
    We therefore split the elevation range into a Side group and a Top group, so that each problem can be evaluated on its own and in combination with the other.
    Matching across the two groups is where the difficulty should arise, because query and gallery images then come from the two different problems, and training on a single group leaves the other one unseen.
    The resulting scenarios give a common basis for comparing ReID methods on their ability to generalize across viewpoint changes.

    \subsection{The Primary ReID Split}
    \label{subsec:primary-split}

    The primary split divides the dataset at the threshold identified in the elevation analysis.
    Side contains images below $30^\circ$, Top contains images at or above $30^\circ$, and Top-Side is their union.
    The training configurations are Side only, Top only, and Top-Side, always using the same $500$ training identities.

    For the $500$ test identities, we define the five query $\rightarrow$ gallery scenarios in \cref{tab:protocols}.
    The scenario describes the query and gallery ranges independently of the training configuration.
    Each of the three training configurations is evaluated on all five scenarios.
    In particular, same-range evaluation means that query and gallery share an elevation range, even when the model was trained on another range.
    The cross-view scenarios test whether the model can match an individual despite the change in perspective, while the mixed scenario includes both same-range and cross-view matches.

    \begin{table}[htbp]
        \centering
        \begin{tabular}{lll}
            \toprule
            \textbf{Setting} & \textbf{Query} & \textbf{Gallery} \\
            \midrule
            Same-range       & Side           & Side             \\
            Same-range       & Top            & Top              \\
            Cross-view       & Side           & Top              \\
            Cross-view       & Top            & Side             \\
            Mixed            & Top-Side       & Top-Side         \\
            \bottomrule
        \end{tabular}
        \caption{
            Five query-to-gallery scenarios of the MOO benchmark.
            Two scenarios are same-range, two are cross-view, and one is mixed.
        }
        \label{tab:protocols}
    \end{table}

    These scenarios also correspond to real monitoring setups.
    For example, a ceiling-mounted fisheye camera captures top views of animals directly below it and more lateral views of animals farther away.
    The benchmark can be used to compare models for deployment in such a system.

    \subsection{Same-View Evaluation}
    \label{subsec:same-view}

    \Cref{tab:benchmark} reports mAP and rank-1 accuracy for the three training configurations across all five query-gallery scenarios.
    We first examine the two same-range scenarios, Side $\rightarrow$ Side and Top $\rightarrow$ Top.
    Training on top views reaches $99.6\%$ mAP on Top $\rightarrow$ Top when all images have an elevation of at least $30^\circ$.
    This result supports the threshold found on the atomic partitions, where views between $30^\circ$ and $90^\circ$ contain enough overlapping content for reliable matching.
    The same does not hold for Side $\rightarrow$ Side, which reaches at most $42.2\%$ mAP because each image shows only one flank.
    Rank-1 accuracy reaches $99.7\%$ for side-only training on Side $\rightarrow$ Side, compared with $100\%$ for top-only training on Top $\rightarrow$ Top.
    Thus, retrieving one correct match remains reliable in both settings, while the much lower side-view mAP shows that retrieving all images of an identity is harder.

    \subsection{Cross-View and Mixed Evaluation}
    \label{subsec:cross-view}

    The cross-view scenarios require matching top and side images, as in AG-ReID, and remain difficult even under controlled synthetic conditions.
    Among the three scenarios that require matching across elevations, the best result is $52.5\%$ mAP, obtained by training and evaluating on Top-Side.
    The benefit of covering both elevations is also clear on Side $\rightarrow$ Top, where training on both views reaches $41.6\%$, compared with $13.0\%$ for side-only training and $25.0\%$ for top-only training.

    Overall, training on both viewpoints is best whenever query and gallery sets mix top and side images, as the Side $\rightarrow$ Top, Top $\rightarrow$ Side, and Top-Side $\rightarrow$ Top-Side results show.
    When comparing the two single-view training configurations within each cross-view scenario, training on the query viewpoint gives better rank-1 accuracy, while top-only training gives better mAP in both scenarios.
    This matters when only the first retrieved candidate is used.
    Training on both viewpoints slightly reduces same-range performance, as Side $\rightarrow$ Side falls from $42.2\%$ to $37.0\%$, and Top $\rightarrow$ Top from $99.6\%$ to $93.5\%$.
    The benchmark thus recovers the same trade-off as the cumulative-elevation experiments, with the model exchanging specialization for coverage.

    \begin{table}[htbp]
        \centering
        \resizebox{\linewidth}{!}{%
            \begin{tabular}{l cc cc cc cc cc}
                \toprule

                & \multicolumn{2}{c}{\textbf{Side $\rightarrow$ Side}}
                & \multicolumn{2}{c}{\textbf{Top $\rightarrow$ Top}}
                & \multicolumn{2}{c}{\textbf{Top $\rightarrow$ Side}}
                & \multicolumn{2}{c}{\textbf{Side $\rightarrow$ Top}}
                & \multicolumn{2}{c}{\textbf{Top-Side $\rightarrow$ Top-Side}} \\

                \cmidrule(lr){2-3} \cmidrule(lr){4-5} \cmidrule(lr){6-7} \cmidrule(lr){8-9} \cmidrule(lr){10-11}

                \textbf{Training}
                & \textbf{mAP} & \textbf{R-1}
                & \textbf{mAP} & \textbf{R-1}
                & \textbf{mAP} & \textbf{R-1}
                & \textbf{mAP} & \textbf{R-1}
                & \textbf{mAP} & \textbf{R-1} \\

                \midrule

                Train Side
                & \textbf{42.2} & \textbf{99.7}
                & 16.5 & 94.4
                & 13.1 & 35.6
                & 13.0 & 75.5
                & 20.6 & 97.4 \\

                Train Top
                & 11.0 & 73.4
                & \textbf{99.6} & \textbf{100}
                & 22.0 & 94.7
                & 25.0 & 29.7
                & 39.4 & 87.2 \\

                Train Top-Side
                & 37.0 & 99.3
                & 93.5 & \textbf{100}
                & \textbf{39.4} & \textbf{96.2}
                & \textbf{41.6} & \textbf{79.7}
                & \textbf{52.5} & \textbf{99.7} \\

                \bottomrule
            \end{tabular}%
        }
        \caption{
            MOO retrieval after training on Side views, Top views, or both.
            Arrows run from query to gallery. Side denotes elevations below $30^\circ$ and Top denotes elevations at or above $30^\circ$, with all azimuths retained.
            mAP and rank-1 accuracy (R-1) are percentages.
            Bold marks the best training configuration for each metric and retrieval scenario.
        }
        \label{tab:benchmark}
    \end{table}

    \section{Transfer to Real Cattle Datasets}
    \label{sec:transfer}

    The analysis and the benchmark are both conducted on rendered images, which may not reflect the challenges of real-world data.
    To verify that MOO is useful beyond the synthetic setting, we test whether it can provide pretraining for real-world datasets.
    We use four real datasets, namely FriesianCattle2015 (FC15) \citep{andrewAutomaticIndividualHolstein2016}, FriesianCattle2017 (FC17) and AerialCattle2017 (AC17) \citep{andrewVisualLocalisationIndividual2017}, and Cows2021 (C21) \citep{gaoSelfSupervisionVideoIdentification2021b}.
    For each dataset, we compare the standard ImageNet-21k initialization with an additional MOO pretraining stage using either all views or top views only.
    Because the architecture and target-dataset training remain fixed, the comparison measures the effect of adding MOO pretraining.

\begin{figure}[htbp]
        \centering
        \includegraphics[width=0.6\linewidth]{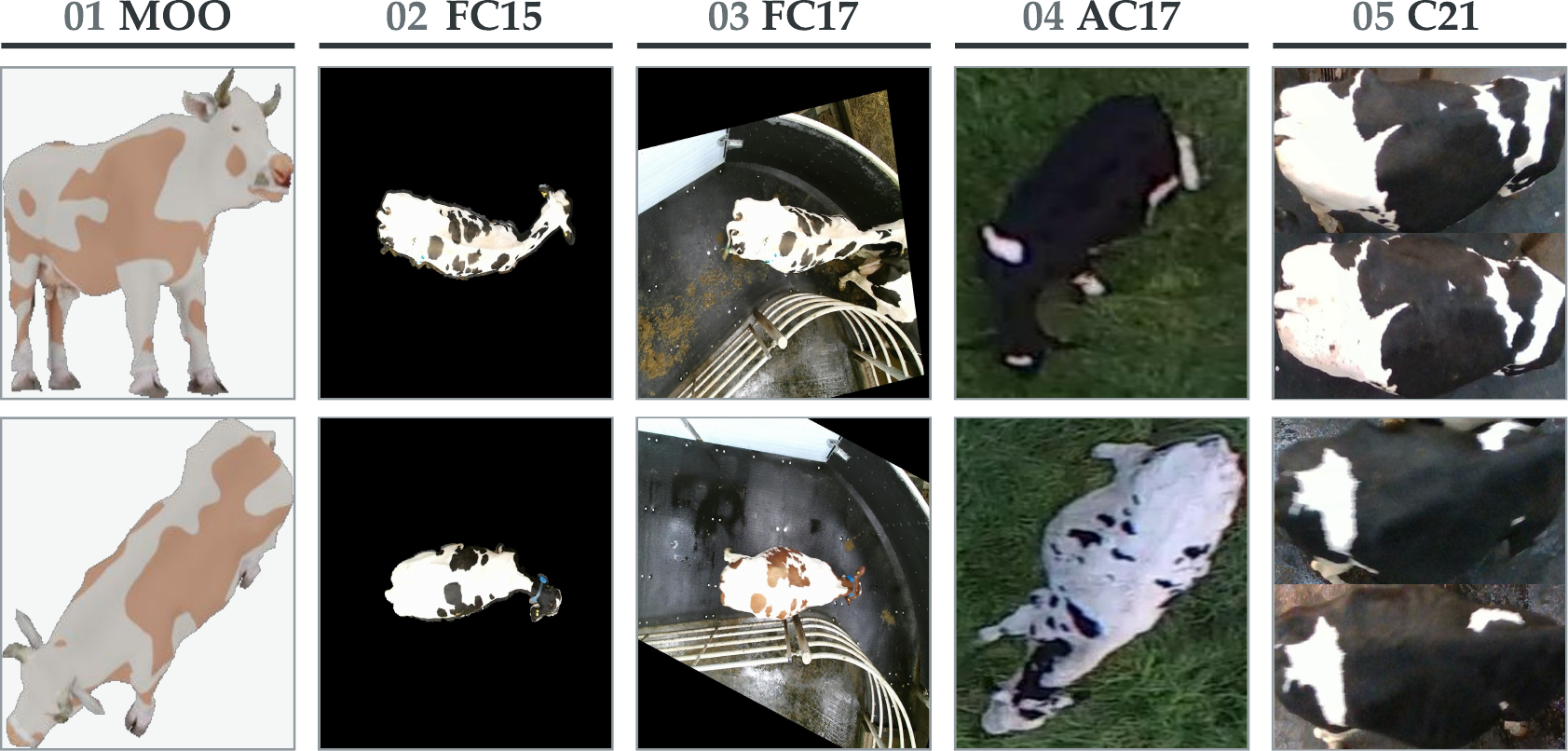}
        \caption{
            Synthetic source images from MOO (01) alongside real target images from FriesianCattle2015 (FC15) \citep{andrewAutomaticIndividualHolstein2016} (02), FriesianCattle2017 (FC17) (03), AerialCattle2017 (AC17) \citep{andrewVisualLocalisationIndividual2017} (04), and Cows2021 (C21) \citep{gaoSelfSupervisionVideoIdentification2021b} (05).
            Each C21 panel contains two adjacent crops.
        }
        \label{fig:transfer-examples}
    \end{figure}

    \Cref{fig:transfer-examples} illustrates the acquisition differences between the synthetic source (01) and the real target datasets (02--05).

    \Cref{tab:real} reports the results of these initializations on each dataset.
    In the zero-shot setting, top-view MOO pretraining outperforms ImageNet-21k on every dataset, while all-view pretraining falls below it on FC17.
    The viewpoint annotations also let us select pretraining images that match the target dataset.
    On Cows2021, top-view pretraining reaches $32.1\%$ mAP, compared with $13.4\%$ for all views and $9.4\%$ for the baseline.
    Thus, datasets dominated by specific views can benefit from pretraining on the relevant subset of MOO.

    The supervised setting then asks whether this advantage holds once the model is also trained on the target dataset.
    Here, MOO initialization improves performance on FC15 and AC17, is comparable on C21, and reduces it on FC17.
    FC17 images contain frequent occlusions and multiple animals per frame, a domain gap that may explain the lower transfer performance.
    Top-view pretraining gives the best results on most of these top-down targets.
    Synthetic data can thus complement real data that is hard to collect.

    \begin{table}[htbp]
        \centering
        \resizebox{\linewidth}{!}{%
            \begin{tabular}{l cc cc cc cc}
                \toprule

                & \multicolumn{2}{c}{\textbf{FC15}} & \multicolumn{2}{c}{\textbf{FC17}} & \multicolumn{2}{c}{\textbf{AC17}} & \multicolumn{2}{c}{\textbf{C21}} \\

                \cmidrule(lr){2-3}
                \cmidrule(lr){4-5}
                \cmidrule(lr){6-7}
                \cmidrule(lr){8-9}

                \textbf{Initialization $\rightarrow$ Pretraining} & \textbf{mAP}  & \textbf{R-1}  & \textbf{mAP}  & \textbf{R-1}  & \textbf{mAP}  & \textbf{R-1} & \textbf{mAP} & \textbf{R-1} \\

                \midrule
                \multicolumn{9}{c}{\emph{Zero-shot (direct transfer)}} \\
                \midrule

                ImageNet-21k (baseline)                           & 51.1          & 46.8          & 45.8          & 60.0          & 55.5          & 79.6          & 9.4           & 32.4          \\
                ImageNet $\rightarrow$ MOO (All)                  & 59.2          & 56.0          & 40.1          & 57.1          & \textbf{67.5} & \textbf{88.0} & 13.4          & 41.2          \\
                ImageNet $\rightarrow$ MOO (Top)                  & \textbf{63.4} & \textbf{56.7} & \textbf{53.9} & \textbf{63.3} & 65.3          & 84.4          & \textbf{32.1} & \textbf{67.4} \\

                \midrule
                \multicolumn{9}{c}{\emph{Supervised (trained on target)}} \\
                \midrule

                ImageNet-21k (baseline)                           & 73.7          & 79.6          & \textbf{90.0} & \textbf{89.5} & 81.8          & 91.2          & 94.0          & 98.0          \\
                ImageNet $\rightarrow$ MOO (All)                  & 89.2          & 94.4          & 83.9          & 86.8          & 82.4          & 91.7          & 94.2          & 98.0          \\
                ImageNet $\rightarrow$ MOO (Top)                  & \textbf{91.1} & \textbf{96.3} & 84.2          & 86.5          & \textbf{89.0} & \textbf{92.8} & \textbf{94.4} & \textbf{98.3} \\

                \bottomrule
            \end{tabular}%
        }
        \caption{
            Transfer from ImageNet-21k initialization, with or without additional MOO pretraining on all views or Top views ($\theta\geq30^\circ$).
            Zero-shot evaluation uses no target-dataset training, and supervised evaluation fine-tunes on the target dataset.
            FC15, FC17, AC17, and C21 denote FriesianCattle2015, FriesianCattle2017, AerialCattle2017, and Cows2021.
            mAP and rank-1 accuracy (R-1) are percentages. Bold marks the best value per column within each block.
        }
        \label{tab:real}
    \end{table}

    \section{Limitations}
    \label{sec:limitations}

    \paragraph{Cross-view evaluation coverage}
    The atomic experiments in \cref{sec:atomic} use query and gallery images from the same elevation or azimuth partition.
    They measure generalization from training viewpoints to test viewpoints, but do not test cross-view retrieval between distinct query and gallery partitions.
    The benchmark in \cref{subsec:primary-split} introduces cross-view retrieval between the broader Top and Side groups, but does not cover every pair of atomic partitions.
    Consequently, it does not establish how retrieval performance varies with a specific query-gallery elevation or azimuth gap.
    Because the possible combinations of partitions are too numerous to evaluate exhaustively, we restricted cross-view retrieval to the Top and Side groups, which correspond to the side and top cameras of real monitoring setups.

    \paragraph{Interactions with other appearance factors}
    The controlled rendering conditions also limit the scope of the conclusions.
    Geometry, pose, lighting, and background remain fixed, so the experiments do not measure how each of these factors interacts with viewpoint changes.
    The real-data transfer experiments assess the usefulness of MOO pretraining, but several appearance and acquisition factors change together and cannot be disentangled by that comparison.
    We deliberately chose a simple rendering to study the effect of viewpoint without the influence of external factors.
    This work is therefore not a study of how geometry, pose, lighting, or background affect ReID.

    \paragraph{Comparison of methods}
    The viewpoint experiments use a single baseline pipeline rather than comparing a range of state-of-the-art ReID methods.
    The measured asymmetry, elevation threshold, and performance gaps therefore characterize this pipeline on MOO, and their magnitude may vary when other architectures or training objectives are used.
    The benchmark enables such comparisons.
    We deliberately leave the evaluation of each method outside the scope of this work, which focuses on measuring the effect of viewpoint.

    \section{Conclusion}
    \label{sec:conclusion}

    We introduced MOO, a synthetic dataset that makes viewpoint effects in animal ReID measurable through $1{,}000$ cattle identities rendered from $128$ systematically sampled viewpoints with exact azimuth and elevation annotations.
    By controlling every source of appearance variation except the coat pattern and viewpoint, it connects angular changes directly to identification performance.

    Our analysis measures the effects of elevation and azimuth.
    Elevation generalization on cattle is asymmetric, because a model trained on high elevations transfers downward better than a model trained on low elevations transfers upward.
    Below a $30^\circ$ elevation threshold, limited overlap between visible surfaces makes retrieval more difficult.
    Higher camera elevations and training data that covers the deployment viewpoints are therefore preferable in this controlled setting.
    Lateral views also perform about $20$ mAP points better than frontal and rear ones on average over the eight test partitions, which we attribute to the larger amount of coat pattern they expose.
    They should be prioritized in both query and gallery sets.

    The MOO benchmark makes viewpoint a central difficulty with disjoint training and test identities, three training configurations, and five query-gallery scenarios that control access to top and side views.
    Since geometry, pose, lighting, and background stay fixed, viewpoint drives the difficulty of these comparisons.
    With top-only training, the baseline reaches $99.6\%$ mAP on Top $\rightarrow$ Top.
    Even with training on both view ranges, it reaches only $39.4\%$ on Top $\rightarrow$ Side and $41.6\%$ on Side $\rightarrow$ Top.
    These cross-view scenarios provide challenging evaluation conditions for methods designed for viewpoint changes.

    The transfer experiments show that MOO pretraining also helps on real datasets.
    After supervised training, it improves performance on two of the four real datasets, is comparable on a third, and reduces it on the fourth.
    Top-view pretraining gives the best results on most of these top-down targets.
    The same annotations thus support both controlled analysis and targeted pretraining.

    Together, these findings isolate and quantify how viewpoint changes affect the ability of a ReID system to match individuals.
    In this controlled cattle setting, matching depends only on the overlap of visible identity cues and the ability to generalize from training viewpoints to test viewpoints.
    A model trained on one viewpoint does not generalize easily to another, and even a model trained on both top and side views struggles to match images across them.
    Viewpoint thus remains a difficulty of its own for animal ReID, even without the occlusion, illumination, and background changes of real data.

    \section*{Acknowledgements}
    This publication was made possible by the use of the FactoryIA supercomputer, financially supported by the Ile-De-France Regional Council.

    \bibliographystyle{elsarticle-harv}
    \bibliography{references}

\end{document}